\documentclass{article}

 \usepackage[preprint]{neurips_2025}

\usepackage[utf8]{inputenc} %
\usepackage[T1]{fontenc}    %
\usepackage{url}            %
\usepackage{booktabs}       %
\usepackage{amsfonts}       %
\usepackage{nicefrac}       %
\usepackage{microtype}      %
\usepackage{xcolor}         %
\usepackage{algorithm}
\usepackage{algorithmic}
\usepackage{amsmath}
\usepackage{multirow}
\usepackage{graphicx} 
\usepackage{subcaption}
\usepackage{wrapfig}  %

\usepackage[most]{tcolorbox}
\usepackage{fancyvrb,fvextra}
\usepackage[scaled=0.8]{FiraMono}

\usepackage{enumitem}     %
\usepackage{setspace}     %
\usepackage{hyperref}       %

\newtcbox{\inlinebox}[1][]{enhanced,
 box align=base,
 nobeforeafter,
 colback=gray!30,
 colframe=gray!30,
 size=small,
 left=0pt,
 right=0pt,
 boxsep=1pt,
 #1}

\title{Self-Abstraction from Grounded Experience for Plan-Guided Policy Refinement}

\author{
Hiroaki Hayashi\thanks{Core Contributor}\footnotemark[1] \footnotemark[2]\,,
    Bo Pang\footnotemark[1]\,,
    Wenting Zhao\footnotemark[1]\,,
    Ye Liu\footnotemark[1]\,,
    Akash Gokul\footnotemark[1]\\ 
    \textbf{Srijan Bansal, Caiming Xiong, Semih Yavuz, Yingbo Zhou\thanks{Corresponding Author}\footnotemark[2]} \\
    Salesforce AI Research \\
    {\tt \{hiroakihayashi, yingbo.zhou\}@salesforce.com} \\
}

\begin{document}

\maketitle

\begin{abstract}
Large language model (LLM) based agents are increasingly used to tackle software engineering tasks that require multi-step reasoning and code modification, demonstrating promising yet limited performance. 
However, most existing LLM agents typically operate within static execution frameworks, lacking a principled mechanism to learn and self-improve from their own experience and past rollouts.
As a result, their performance remains bounded by the initial framework design and the underlying LLM’s capabilities.
We propose \textbf{S}elf-\textbf{A}bstraction from \textbf{G}rounded \textbf{E}xperience (\textbf{SAGE}), a framework that enables agents to learn from their own task executions and refine their behavior through self-abstraction.
After an initial rollout, the agent induces a concise plan abstraction from its grounded experience, distilling key steps, dependencies, and constraints.
This learned abstraction is then fed back as contextual guidance, refining the agent’s policy and supporting more structured, informed subsequent executions.
Empirically, SAGE delivers consistent performance gains across diverse LLM backbones and agent architectures.
Notably, it yields a 7.2\% relative performance improvement over the strong Mini-SWE-Agent baseline when paired with the GPT-5 (high) backbone.
SAGE further achieves strong overall performance on SWE-Bench Verified benchmark, reaching 73.2\% and 74\% Pass@1 resolve rates with the Mini-SWE-Agent and OpenHands CodeAct agent framework, respectively.

\end{abstract}

\section{Introduction}

Large language model (LLM) based agents have made rapid progress on realistic software-engineering tasks: most notably bug fixing \cite{jimenez2024swebench,antoniades2025swesearch,liu-etal-2025-repodebug,zhang2024diversity}, repository-level code generation \cite{zhang2023repocoder,liu2023repobench,li2024evocodebench}, and refactoring \cite{kovacic2025refactoring}.
Recent work shows that LLM agents' performance can be greatly improved with 1) reflective/self-improvement loops that critique and revise an agent’s own outputs without weight updates~\cite{shinn2023reflexion, madaan2023self}; 2) richer agent-computer interfaces that let models operate development environments end-to-end~\cite{yang2024swe}; and 3) test-time scaling with search~\cite{zhou2023language,li2024codetree}, and ensembles~\cite{traeresearchteam2025traeagent}. However, most of these works still approach each task ``from scratch,'' expecting models to solve the task in their first rollout, thus limiting the model's ability to uncover relevant insight into the problem and execution environment. 

Building on these advances, a central open question remains: \emph{How should agents learn and improve from their own grounded experience at test-time}? Prior works focused on reflecting upon past steps within a single rollout, which implicitly assumes the agent can solve the task in its first attempt, an assumption that fails to hold in complex, long-horizon tasks. Instead, it is crucial that agents are able to perform multiple rollouts and effectively reflect upon these rollouts in order to solve the task, akin to a human taking multiple attempts at a problem and reflecting upon those attempts before finally solving it.

\begin{figure}[t!]
    \centering
    \includegraphics[width=\linewidth]{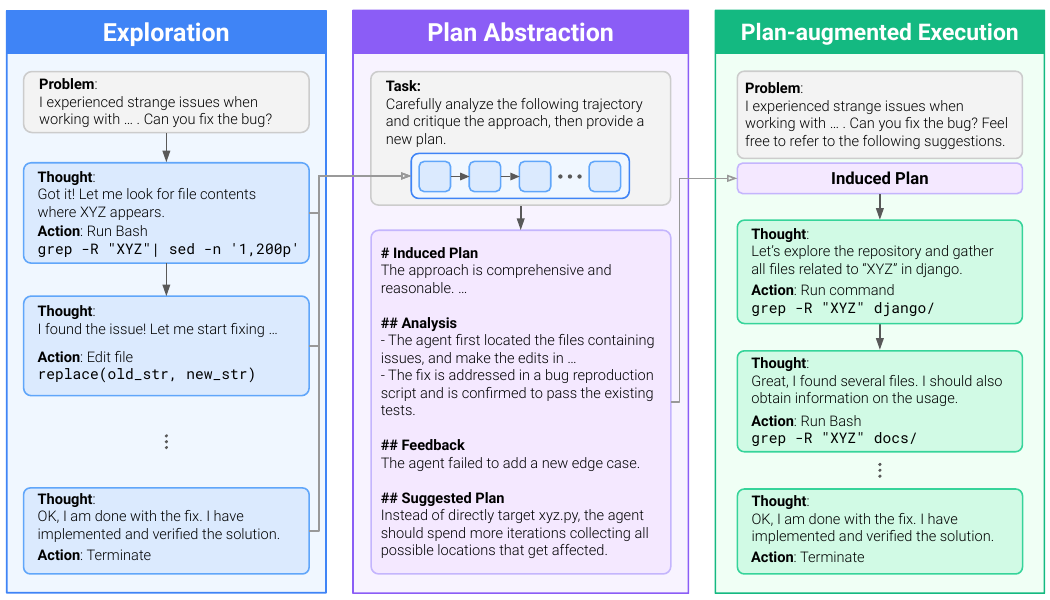}
    \caption{SAGE consists of three stages. (1) Exploration: the agent attempts to finish the given task. (2) Plan Abstraction: an agent critically analyze and suggest a high-level plan based on the exploration trajectory. (3) Plan-augmented execution: an agent attempts to finish the task again with access to the high-level plan. Each stage can be instantiated with same or different LLM backbones \& agents. We match the settings for Exploration and Plan-augmented execution throughout the experiments, while investigating different LLM backbones for Plan abstraction.}
    \label{fig:sage_figure}
\end{figure}

An agent’s previous trajectories encode rich signals about the environment, reasoning process, and past actions. However, directly feeding the raw trajectory back into the LLM is inefficient: these trajectories are often long, noisy, and exceed context-window limits, making it difficult for the model to identify what truly matters. We address this challenge through \textbf{S}elf-\textbf{A}bstraction from \textbf{G}rounded \textbf{E}xperience (\textbf{SAGE}), a structured cycle of rollout, reflection, and re-execution that transforms experience into actionable guidance, as illustrated in Figure~\ref{fig:sage_figure}.

To concretely study this process, we use software-engineering (SWE) agents on the bug-fixing task as our primary testbed.
SAGE operates in three stages: \emph{Exploration}, where the agent interacts with the environment to identify and fix a bug, producing a detailed but unstructured trajectory; \emph{Plan Abstraction}, where it distills this trajectory into a concise, high-level plan summarizing key steps, rationales, and limitations; and \emph{Plan-Augmented Execution}, where it re-executes the task using the induced plan as additional context—acting more strategically and addressing missed edge cases. This abstraction-guided cycle enables the agent to learn from grounded experience, transforming a single rollout into reusable knowledge for more effective subsequent reasoning.

We empirically validate our proposed method on SWE-Bench Verified~\cite{jimenez2024swebench,chowdhury2024swebenchverified}, a program repair benchmark of 500 human-verified GitHub issues. We demonstrate that SAGE consistently provides gains in the resolution rate across different LLM backbones and agent frameworks. Furthermore, we find that performance improves further by assigning distinct LLMs to individual stages, achieving a 74\% resolution rate. These findings echo the broader value of \emph{self-improvement cycles} seen across the technology industry, where structured reflection and iterative refinement drive sustained innovation and stronger problem-solving capabilities.

The contributions of this work are as following: 
\begin{enumerate}
    \item We propose \textbf{SAGE} — a general test-time adaptation framework that enables an agent to learn and improve from its own grounded experience. SAGE induces a plan abstraction from the agent's trajectory and then conditions the agent’s subsequent execution on this plan, realizing \emph{plan-guided policy refinement}.

    \item We demonstrate its effectiveness through comprehensive experiments across different LLM backbones (e.g., GPT, Claude, Gemini) and agent architectures. Our method shows consistent improvements compared with baselines where plan guidance is unavailable.

\end{enumerate}

\section{Method}
\label{sec:method}

\subsection{Background}
We consider the software engineering agent as an Markov Decision Process (MDP) with tuple $\mathcal{M}$
\begin{equation}
\label{eq:pomdp}
\mathcal{M} = (\mathcal{S},\,\mathcal{A},\,T,\,R,\,\gamma),
\end{equation}
where $\mathcal{S}$ is the state space (\textit{e.g.}, repository content, build/test status, past action and observations, etc.); $\mathcal{A}$ is the action space (\textit{e.g.}, tool calls, code edits, testing); $T(s'\mid s,a) \in \mathcal{S} \times \mathcal{A} \mapsto \mathcal{S}$ is the transition function for the underlying environment; $R(s,a) \in \mathcal{S} \times \mathcal{A} \mapsto \mathbb{R}$ is a reward function; and $\gamma\in[0,1)$ is a discount factor. A trajectory from an agent following a policy $\pi$ up to time $t$ is defined as $\tau_\pi^t=(s_1, a_1, r_1, s_2, a_2, r_2, \ldots, s_t, a_t, r_t)$.

\subsection{Self-Abstraction from Grounded Experience}

Self-Abstraction from Grounded Experience (SAGE) instantiates three agent roles -- exploration, plan abstraction, and plan-augmented execution, which act over the MDP process $\mathcal{M}$. 
\paragraph{Exploration} First, given the task context (initial state) $s_0$, the exploration agent $\mathcal{A}_{\theta}$ executes the task until time step $T_E$ (or until task completion) under its initial policy $\pi_\theta$ to gather experience:
\begin{equation}
\label{eq:explore}
a_t \sim \pi_\theta(\cdot\mid S_t),\quad s_{t+1}\sim T(\cdot\mid s_t,a_t),
\end{equation}
producing a trajectory $\tau_{\theta}^{T_E} = (s_1,a_1,r_1\ldots,s_{T_E},a_{T_E},r_{t_E})$ and (optionally) return $G = \sum_{t=1}^T \gamma^{t-1} r_t$. The trajectory serves as a source for grounded experience of the agent in the environment when performing the given task. Note that the agent $\mathcal{A}_{\theta}$ is the agent one would like to improve based on the experience. It can be any agent as long as the agent can function in the target environment.

\paragraph{Plan Abstraction} 
After the exploration, the plan abstraction agent $\mathcal{P}_{\phi}$ induces a high-level plan $\psi$ from the previous experience:
\begin{equation}
\label{eq:plan_induce}
\psi \sim \mathcal{P}_{\phi}(\cdot | \tau_{\theta}^{T_E}).
\end{equation}
The plan is designed to capture key information for subsequent policy refinement. On the one hand, it summarizes high-level information about the agent's prior execution and interactions with the environment. 
This enables the agent to assess its previous strategy and its alignment with the task.
On the other hand, the abstraction is grounded in the past experience with real environment transitions. 
This grounding provides crucial information such that the agent's internal plan and expectations are better aligned with the actual environment when incorporated for later executions.

\paragraph{Plan-augmented Execution} Finally, the plan-augmented agent $\mathcal{A}_{\theta}^{+}$ attempts the same task with a revised policy $\pi_{\theta}^{+}$ that is augmented and refined by the generated plan $\psi$.
\begin{equation}
\label{eq:plan_augment}
    a_t \sim \pi_{\theta}^+(\cdot | s_t, \psi).
\end{equation}
In other words, the agent $\mathcal{A}_{\theta}^{+}$ follows the same agent framework as $\mathcal{A}_{\theta}$ but with a revised policy $\pi_\theta^+$ 

With a raw trajectory consisting of primitive actions being abstracted to a concise plan for more informed subsequent rollouts, we can view this plan as an \textit{option}~\cite{sutton1999between}, defined as a tuple of (a) an initiation set $\mathcal{I}_o$, (b) an intra-option policy $\mu_o$, and (c) a termination condition.
In our setting, $\mathcal{I}_o$ is the terminal state of the exploration agent $\mathcal{A}_\theta$ and $\mu_o$ is the plan-augmented policy $\pi^{+}_\theta$ given the plan as the option $o$.

\paragraph{LLM parameterization} When adapting to existing LLM based agents, the actor policy $\pi_\theta$ is implemented by an LLM that emits a text response which is decoded to an action $a_t$ at time $t$. Let $\mathrm{decode}:\mathcal{W}^*\!\mapsto\mathcal{A}$ be a map between a well-formed token span $w$ to an action. Then
\begin{equation}
\label{eq:llm_policy}
x_t = f(s_t), \qquad w_k \sim p_\theta(\cdot\mid w_{<k},\, x_t), \qquad a_t = \mathrm{decode}(w),
\end{equation}
where $f(\cdot)$ is the observation function that converts states into textual format for LLMs to consume. 
We show an implementation of SAGE using LLM parameterization in Algorithm \ref{alg:sage}.

\begin{algorithm}[t]
\caption{Self-Abstraction from Grounded Experience with LLM parameterization}
\label{alg:sage}
\begin{algorithmic}[1]
\STATE \textbf{Input:} process $\mathcal{M}$; actor LLM $\mathcal{A}_\theta$; planner LLM $\mathcal{P}_\phi$; initial state $s_1$
\STATE \textbf{Exploration}: Roll out $\pi_\theta$ using Eq~\eqref{eq:explore} to obtain $\tau_\theta=(s_{1:T_E},s_{1:T_E},r_{1:T_E})$
\STATE \textbf{Plan Abstraction}: Using $\mathcal{P}_\phi$, produce a plan $\psi$ from $\tau_{\theta}^{T_E}$ per Eq~\eqref{eq:plan_induce}
\STATE \textbf{Execute with Plan}: Form $\pi^{+}_\theta$ via Eq~\eqref{eq:plan_augment} and re-execute to obtain the solution
\end{algorithmic}
\end{algorithm}

\section{Experiments}
\subsection{Implementation Details}
\paragraph{Implementation} All the exploration, plan, and plan-augmented execution agents (i.e. $\mathcal{A}_{\theta}$, $\mathcal{A}^+_\theta$, and $\mathcal{P}_{\phi}$) are implemented using \texttt{mini-swe-agent}~\cite{yang2024swe}, a simple agent framework that allows interactions with the environment only via bash commands (\textit{e.g.}, file modification, command execution).
For the plan agent $\mathcal{P}_{\phi}$, we provide the issue description and the first roll out trajectory as part of the context. All runs have a step limit of 250 and a cost limit of \$3 USD per instance\footnote{Most of the instances terminate at less than 100 steps, well within the budget.}. 
We evaluate various backend LLMs, including GPT-5-mini \& GPT-5~\cite{openai2025gpt5}, Gemini 2.5 Pro~\cite{Gemini2.5Pro2025}, Claude Sonnet 4 and Claude Sonnet 4.5~\cite{anthropic2025sonnet4,anthropic2025sonnet4.5}.
Appendix~\ref{app:prompts} lists the set of prompts.

\paragraph{Evaluation Setup}
We evaluate our experiments using the official \texttt{SWE-bench} evaluation harness on 500 instances from the SWE-bench Verified dataset~\cite{jimenez2024swebench,chowdhury2024swebenchverified}.
During evaluation, we identified several validation issues where the official gold patches were marked incorrect, which led to false negatives.
To mitigate their impact, we adopt the corresponding instances from the more up-to-date \texttt{SWE-bench} dataset repository\footnote{\url{https://huggingface.co/datasets/princeton-nlp/SWE-bench}} that reflects recent fixes (\textit{e.g.}, \texttt{astropy-7606}, which now removes a non-existent test case).
We further resolve remaining gold patch failure cases, such as compatibility issues in \texttt{astropy-\{8707,8872\}} due to deprecated dependencies\footnote{\url{https://huggingface.co/datasets/inweriok/SWE-bench_Verified_gold_fixes}}.
For instances like \texttt{psf\_\_requests-\{1724,1766,1921,2317\}}, which intermittently failed due to transient \texttt{503} errors from \texttt{httpbin.org}, we introduce a retry mechanism to ensure stable evaluation\footnote{We have submitted this fix as a pull request to the official \texttt{SWE-bench} repository.}.
The resolution rate is computed as the number of resolved instances (after applying the fixes above) divided by the total number of evaluated instances (500).

\subsection{Main Results}

\begin{table}[tb]
\centering
\begin{tabular}{lcc}\toprule
\multirow{2}{*}{Model} & \multicolumn{2}{c}{Resolved {[}\%{]}}\\\cmidrule{2-3}
                & Baseline & SAGE \\\midrule
GPT-5-mini (medium) & 58.6 & \textbf{61.8} (\(\uparrow\)3.2) \\
GPT-5 (medium)      & 65.0 & \textbf{67.2} (\(\uparrow\)2.2) \\
GPT-5 (high)        & 66.6 & \textbf{71.4} (\(\uparrow\)4.8) \\
Gemini 2.5 Pro      & 52.4 & \textbf{53.6} (\(\uparrow\)1.2) \\
Claude Sonnet 4     & 62.0 & \textbf{64.0} (\(\uparrow\)2.0) \\
Claude Sonnet 4.5   & 72.0 & \textbf{72.4} (\(\uparrow\)0.4) \\\bottomrule
\end{tabular}
\caption{Resolution rates (\%) on SWE-Bench Verified comparing baseline and SAGE. Numbers in parentheses indicate absolute improvements. All SAGE runs use the same underlying LLM as their baseline counterpart.}
\label{tab:results}
\end{table}

We show the performance comparison between SAGE and its baseline in Table~\ref{tab:results}.
Here, the baseline refers to unmodified \texttt{mini-swe-agent} runs.
Across different model families and sizes, we observe consistent improvements in the resolution rate.
Notably, the improvement is more significant for weaker models with fewer parameters and reasoning effort, hinting that the generated feedback complements the relative lack of reasoning during rollouts.

Prior work  on LLM-as-a-judge shows that LLMs prefer their own generations~\cite{zheng2023judging,panickssery2024llm}.
Plan induction shares similarities with using LLM-as-a-judge as it requires reviewing and evaluating trajectories sampled from the policy.
When the plan-induction model is the same as the policy model, \textit{i.e.} the same LLM, it is possible that such self-bias leads to a biased assessment and thus suboptimal feedback.
To investigate the impact of self-assessment, we also performed experiments that vary both the policy and the plan induction models.
Results are shown in~Table~\ref{tab:results_diff_models}.
We see that self-assessment indeed impacts the performance, where obtaining plans from other models generally results in a higher resolution rate.

\begin{table}[tb]
\centering
\begin{tabular}{llc}
\toprule
Policy Model      & Plan Abstraction Model & Resolved {[}\%{]} \\ \midrule
GPT-5 (high)      & GPT-5 (high)         & 71.4             \\
GPT-5 (high)      & Claude Sonnet 4.5    & 68.2             \\
GPT-5 (high)      & Gemini 2.5 Pro       & 69.6             \\\midrule
Claude Sonnet 4   & Claude Sonnet 4      & 64.0              \\
Claude Sonnet 4   & GPT-5 (high)         & 68.8              \\\midrule
Claude Sonnet 4.5 & Claude Sonnet 4.5    & 72.4              \\
Claude Sonnet 4.5 & GPT-5 (high)         & 73.2              \\ \bottomrule
\end{tabular}
\caption{Comparisons of resolution rates when using different LLMs for plan abstraction.}
\label{tab:results_diff_models}
\end{table}

\subsection{Comparison with Other Methods}
We compare our method with other methods that also incorporate refinement. 
Specifically, we compare against the following methods: (1) Stepwise Reflection -- the agent gets to reflect at each step of the action, explain its rationale and adjust its action if necessary. (2) Episodic Reflection -- the agent judges its own result at the end of the whole rollout to check for any errors.
If the overall trajectory and the resulting patch are not deemed completely correct by the agent, the agent will then reflect, provide feedback and re-execute on the same task with the reflection feedback (\textit{i.e.}, reflection and re-execution are only done on the instances where the reviewing agent deems it necessary). (3) 2x Scale -- for each task the agent is executed twice independently.
We then select the result based on both the oracle and the LLM judge.

\begin{table}[!tb]
    \centering
    \begin{tabular}{lc}
        \toprule
        Method              & Resolved {[}\%{]} \\ \midrule
        Baseline            & 58.6 \\ \midrule
        Stepwise Reflection & 56.0 \\
        Episodic Reflection & 58.2 \\
        2x Scale (LLM Judge)& 59.2\\
        2x Scale (Oracle)   & 64.8 \\ 
        SAGE (Ours)         & \textbf{61.8} \\
        \bottomrule
    \end{tabular}
    \caption{Ablation results on the SWE-Bench Verified subset using mini-swe-agent framework. Stepwise reflection denotes the agent reflects on its action at each time step. Episodic reflection is the setting where the agent gets to reflect and re-do the task after the task is completed. 2x Scale (LLM Judge) and 2x Scale (Oracle) represent the ensemble results from running the agent twice and select the final patch with an LLM judge and oracle, respectively. }
    \label{tab:ablation_gpt5mini}
\end{table}

Both stepwise and episodic reflection do not help the agent's performance. For stepwise reflection, this may be attributed to the long horizon nature of the task, which makes the local reflection difficult to have a positive influence on the final goal. In the case of episodic reflection, the performance may be impacted by the agent's ability to verify its own solution. As can be seen from the 2x Scale LLM judge result, judging the correctness of the result is itself a challenging problem. 

\subsection{SAGE on Tool-rich Agent Framework}
While our results show consistent improvements across model choices, it is also important to validate the approach across different agent frameworks, because the choice of agent frameworks governs the LLM's decision process (\textit{i.e.}, from an observation to an action), which is reflected in the trajectories.
The trajectory would not only look different in every framework, but might also consist of different granularities.
In particular, \texttt{mini-swe-agent}, which we used throughout the experiments, is a barebone agent framework without any complex tools or context management.

To investigate if our method can function similarly on other agent frameworks, we conducted an experiment with OpenHands CodeAct~\cite{wang2025openhands}, a ReAct-style agent framework with an extensive tool set.
For SAGE, we performed minimal reformatting of OpenHands trajectories into \texttt{mini-swe-agent} style, specifically treating all observations as user turns, eliding long command outputs, and excluding OpenHands-specific actions.
To ensure the trajectory follows the exact CodeAct style where a thought always precedes the action, we supplemented the actions without explicit thought sentences with appropriate deterministic templates such as ``I will create a new file at ...''.
Table~\ref{tab:openhands_ablation} shows that SAGE is effective when implemented with OpenHands, where we are able to further improve OpenHands best results on GPT5 from 71.8\% to 74\%.

\begin{table}[!t]
\centering
\begin{tabular}{lcc}\toprule
\multirow{2}{*}{Model} & \multicolumn{2}{c}{Resolved {[}\%{]}}\\\cmidrule{2-3}
                & Baseline & SAGE \\\midrule
Claude Sonnet 4 & 68.4     & \textbf{71.6} (\(\uparrow\)3.2) \\
GPT-5 (high)    & 71.8     & \textbf{74.0} (\(\uparrow\)2.2) \\\bottomrule
\end{tabular}
\caption{Resolution rate on SWE-Bench Verified using OpenHands CodeAct as the agent framework. We observe consistent improvements when employing SAGE.}
\label{tab:openhands_ablation}
\end{table}

\subsection{LLM Judge Experiments}

Different LLM agents tend to perform better on distinct subsets of issues, exhibiting complementary strengths across models. This diversity motivates an ensemble approach that leverages their collective capabilities. In this section, we evaluate the effectiveness of an LLM-as-a-judge framework in selecting the most effective solution among outputs generated by heterogeneous LLM backbones.
Specifically, we begin by constructing a pool of candidate patches produced by multiple heterogeneous agents, each differing in either the actor model or the plan-generation model. 
For this ensemble analysis, we consider five candidate configurations. Using \texttt{GPT-5 (high)} as the plan-generation model, we sample outputs from four actor models: \texttt{Claude 4.5 Sonnet} (73.2\%), \texttt{Claude 4 Sonnet} (68.8\%), \texttt{GPT-5 (high)} itself (71.4\%), and \texttt{GPT-5 (medium)} (67.2\%). Additionally, we include a configuration pairing \texttt{Gemini-2.5-Pro} as the plan-generation model with \texttt{GPT-5 (high)} as the actor model (69.6\%).

\begin{figure}[t]
\centering
\includegraphics[width=0.75\linewidth]{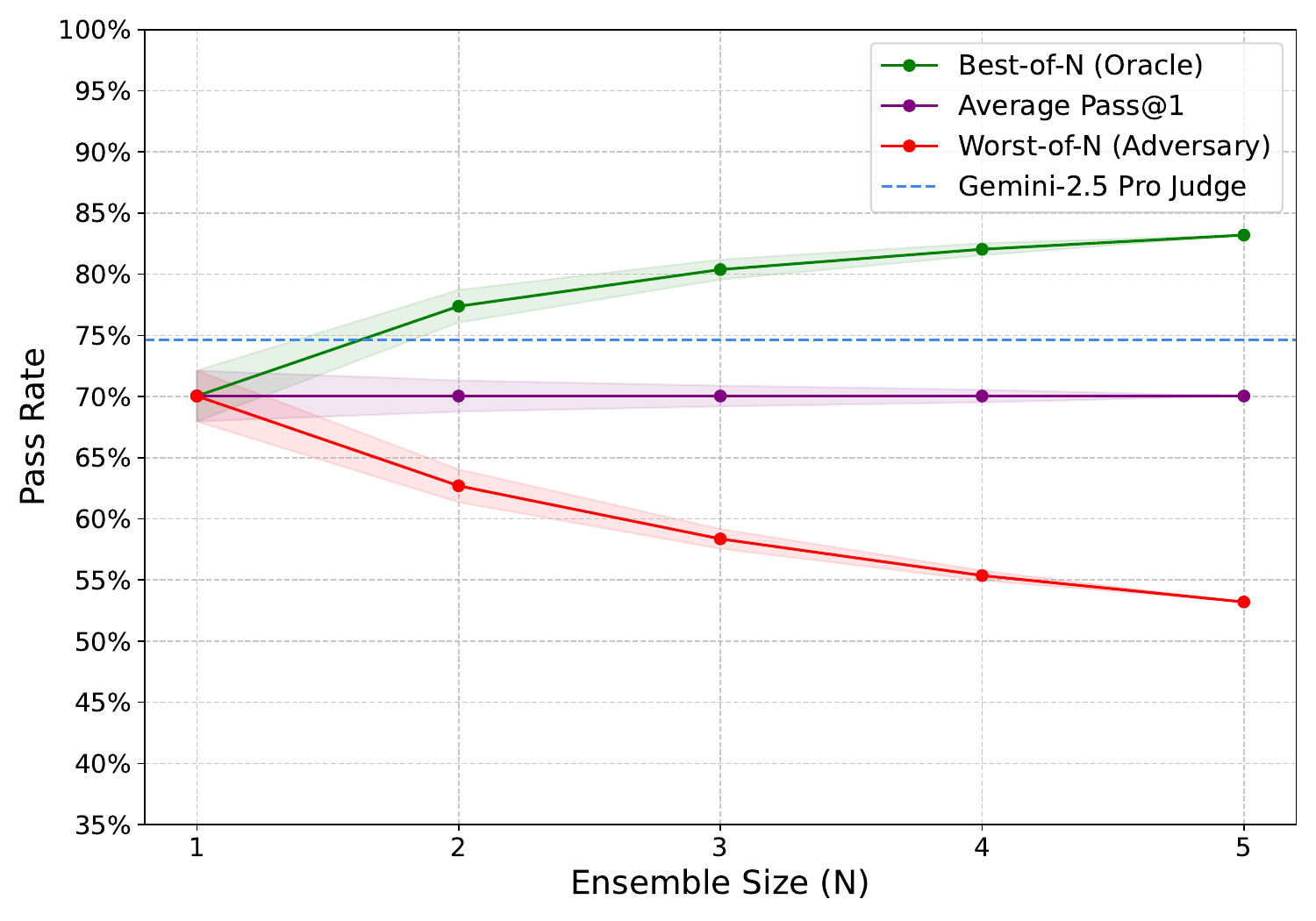}
\caption{\textbf{Influence of candidate diversity on ensemble reasoning performance (Pass@1).} Increasing candidate diversity improves the \textit{Best-of-N (Oracle)} score (up to 83.4\%) while lowering the \textit{Worst-of-N (Adversary)} (to 52.2\%), with mean performance remaining stable around 70\%. The LLM-as-a-judge ensemble achieves 74.6\%, outperforming the average individual model.}

\label{fig:judge}
\end{figure}

As illustrated in Figure~\ref{fig:judge}, this ensemble diversity substantially enhances overall problem coverage. As the number of ensemble candidates increases, the \textit{Best-of-N (Oracle)} performance improves monotonically to 83.4\%, since the likelihood that at least one candidate fully resolves the issue rises with diversity. Conversely, the \textit{Worst-of-N (Adversary)} performance decreases to 52.2\%, reflecting greater variability and disagreement among agents. The mean performance remains relatively stable at around 70\%, indicating a balance between successful and unsuccessful resolutions.

To identify the most effective patch, we employ a frontier LLM (\texttt{Gemini-2.5-Pro}) as a comparative evaluator, following a structured evaluation protocol. The LLM independently examines each candidate, assessing two key dimensions: \textit{correctness} (whether the patch successfully resolves the issue) and \textit{completeness} (whether it addresses all relevant edge cases). It then synthesizes its reasoning to select the best-performing candidate. This ensemble-judge mechanism achieves a final performance of 74.6\%, exceeding the mean performance of individual models. Importantly, the process is computationally lightweight—requiring only a single additional LLM call beyond the generation phase—yet it significantly improves overall patch quality and consistency. In summary, this approach integrates the exploratory breadth of multiple specialized agents with the evaluative precision of an expert LLM, yielding robust and reliable outcomes. %

\section{Discussion}
\label{discussion}

\subsection{Case Study: Plan Induction from Self-Experience}

This section presents a case study illustrating how an agent’s self-experienced patch attempt—originating from its own trial-and-error exploration—can serve as the foundation for plan induction without reliance on any external verifier feedback. It will mainly cover the following stages based on the issue described in Figure \ref{fig:case_study_issue}:
(1) identifying the root cause of the issue in the original repository,
(2) introspecting on the failed patch attempt to induce a reasoning-level plan,
(3) demonstrating how this induced plan re-emerges as explicit structure in the final, correct patch, and
(4) analyzing how individual reflective elements map onto code-level improvements.

\subsubsection{Plan Induction through Introspective Reflection}

After completing its initial exploration, the agent enters an introspective phase in which it revisits its own reasoning traces to derive structured insights.
This process unfolds along three conceptual pillars: \textbf{Analysis} (the intended constraint or invariant the agent aimed to enforce), \textbf{Feedback} (the misconceptions or oversights that arose during execution), and \textbf{Induced Plan} (the reformulation of those insights into a refined strategy for subsequent repair).

\begin{figure}[!t]
\centering

\begin{subfigure}{0.48\textwidth}
    \centering
    \includegraphics[width=\linewidth]{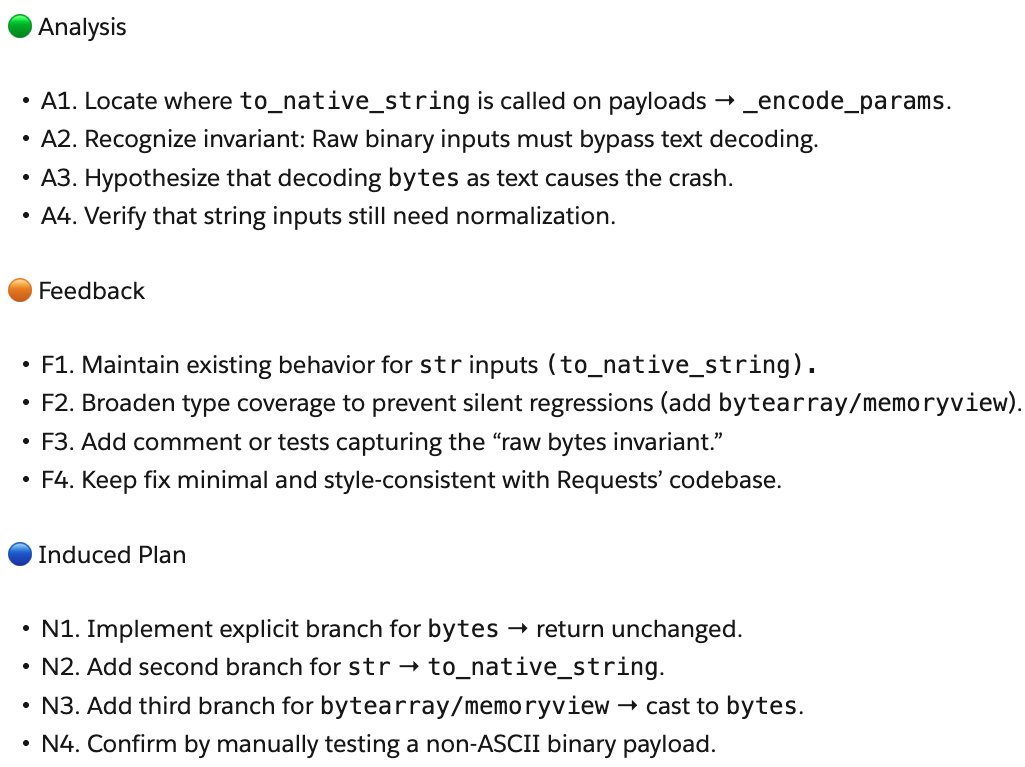}
    \caption{SAGE plan induction from self-experience.}
    \label{fig:case_study_plan}
\end{subfigure}
\hfill
\begin{subfigure}{0.48\textwidth}
    \centering
    \includegraphics[width=\linewidth]{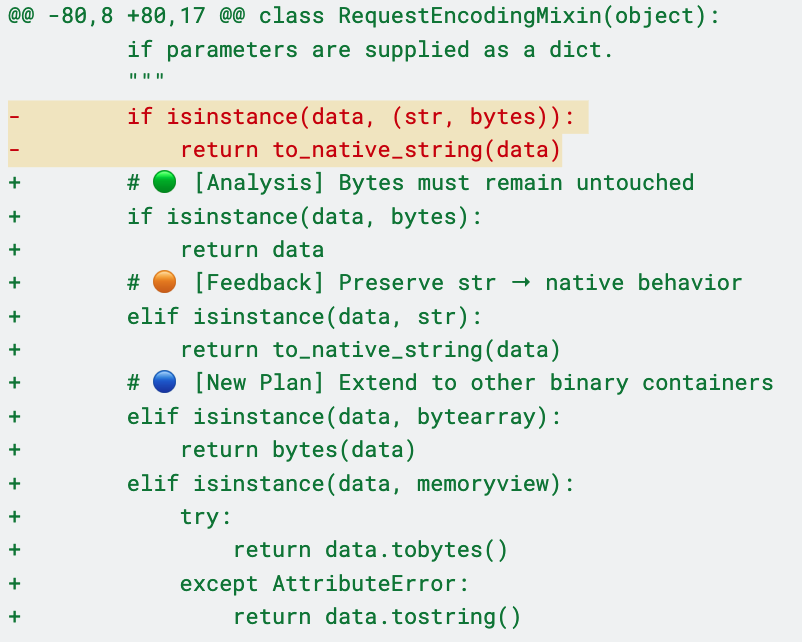}
    \caption{Plan manifestation in the SAGE patch.}
    \label{fig:case_study_attribution}
\end{subfigure}

\caption{Illustration of introspective plan induction and its downstream realization in code based on a real example from SWE Bench Verified improved by SAGE approach.}
\label{fig:case_study}
\end{figure}

Through this decomposition, the agent transforms raw experiential data into a structured plan representation, effectively converting an unstructured exploratory episode into a reusable reasoning artifact. Figure~\ref{fig:case_study} illustrates both (a) the internal plan induction process and (b) how the resulting plan components subsequently materialize in the final patch.

\begin{figure}[t]
\centering

\begin{minipage}{0.95\textwidth}
    \begin{tcolorbox}[colback=gray!3, colframe=gray!40, title={\textbf{Issue description (psf\_\_requests-2931)}}]
        \small
        \textbf{Sending a binary payload such as}
        
        \medskip
        \noindent
        \verb|requests.put(url, data=b"\xc3\xb6\xc3\xb6\xc3\xb6")|
        
        \medskip
        \noindent
        failed in Requests~2.9, although it worked in~2.8.1. 
        The culprit was \texttt{\_encode\_params}, which always called 
        \texttt{to\_native\_string(data)} — decoding raw bytes as ASCII and triggering a \texttt{UnicodeDecodeError}.
        
        \medskip
        \noindent
        \textbf{Goal:} Preserve raw bytes exactly as sent, while still converting text types safely.
    \end{tcolorbox}
    \caption{Issue Description (\textbf{psf\_\_requests-2931}) used in the rest of Case Study.}
    \label{fig:case_study_issue}
\end{minipage}

\end{figure}

The induced plan encapsulates the agent’s distilled reasoning, abstracting the central invariant—e.g., ``avoid decoding raw bytes''--into a reproducible strategy that generalizes to analogous failure modes across different codebases.

\subsubsection{SAGE Patch and Attribution of Reflective Pillars}
Once the reflective plan has been induced, the SAGE (Self-Augmented Guided Exploration) framework leverages it to guide the agent’s subsequent decision-making trajectory.
Rather than resuming exploration through unstructured trial-and-error, the agent now operates under explicit contextual scaffolds derived from its previously surfaced Analysis, Feedback, and Induced Plan components.
This transition transforms exploration into purposeful self-guided repair, in which code edits are informed by the agent’s own learned reasoning patterns.

The final patch, depicted in Figure~\ref{fig:case_study_attribution}, demonstrates how these reflective elements concretely manifest in code.
Specifically, the Analysis component identifies the invariant that raw binary data should bypass encoding; the Feedback component refines the type-handling logic to preserve the legacy string path; and the Induced Plan generalizes the fix to cover all byte-like container types.
Together, these aligned elements illustrate how introspective plan induction enables the agent to internalize and reapply its own reasoning traces, producing patches that are both functionally correct and structurally interpretable.

\subsection{Plan Attribution Analysis}

\begin{figure}[ht]
\centering
\begin{subfigure}{0.6\textwidth}
\centering
\includegraphics[width=\linewidth]{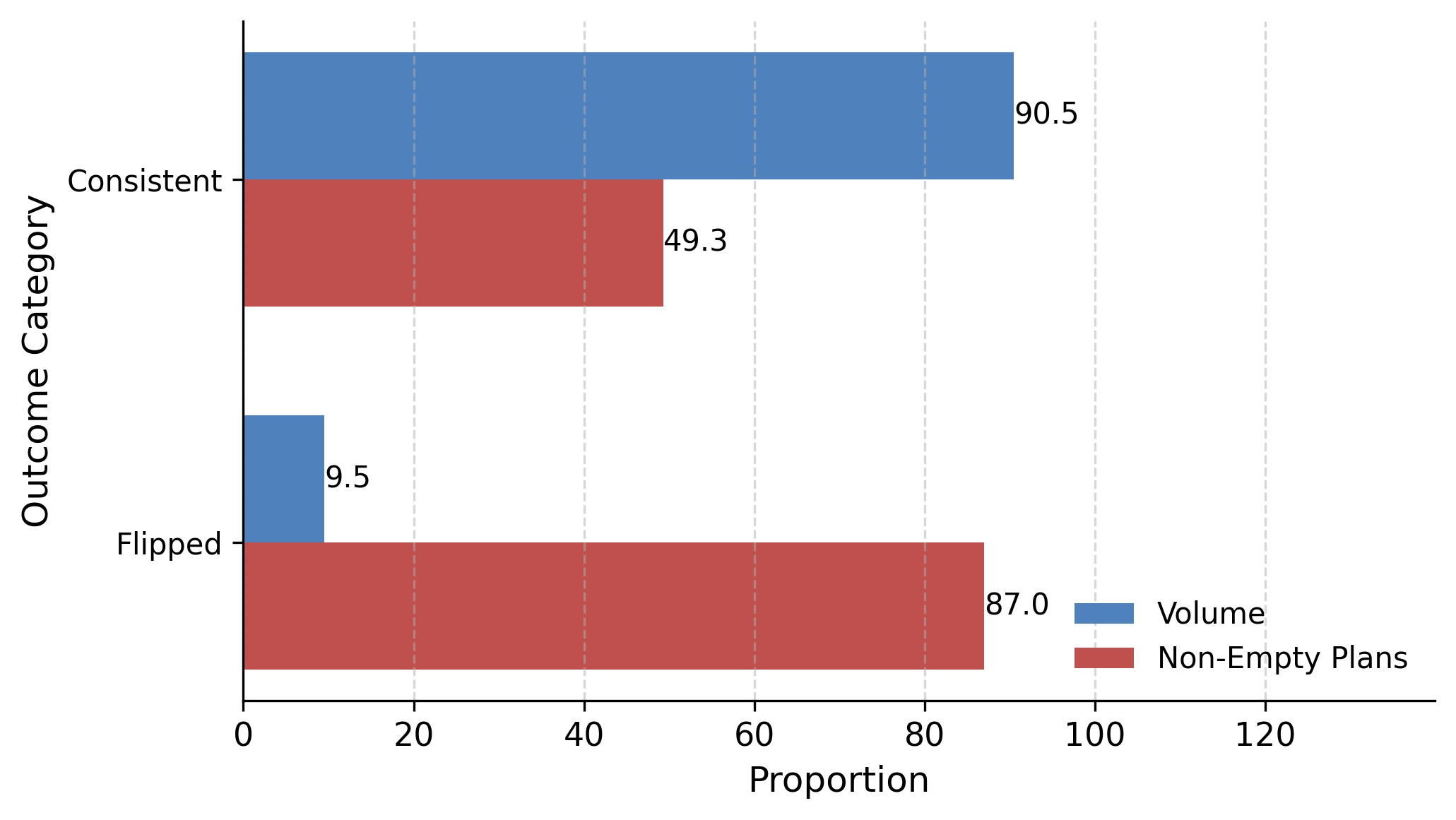}
\caption{Association between outcome changes and non-empty set of plan elements attributed to the code lines in the SAGE patch. It demonstrates a clear trend of identifying traces of the SAGE plan more often when the outcome is flipped.}
\label{fig:plan_attribution_rate}
\end{subfigure}
\hfill
\begin{subfigure}{0.34\textwidth}
\centering
\includegraphics[width=\linewidth]{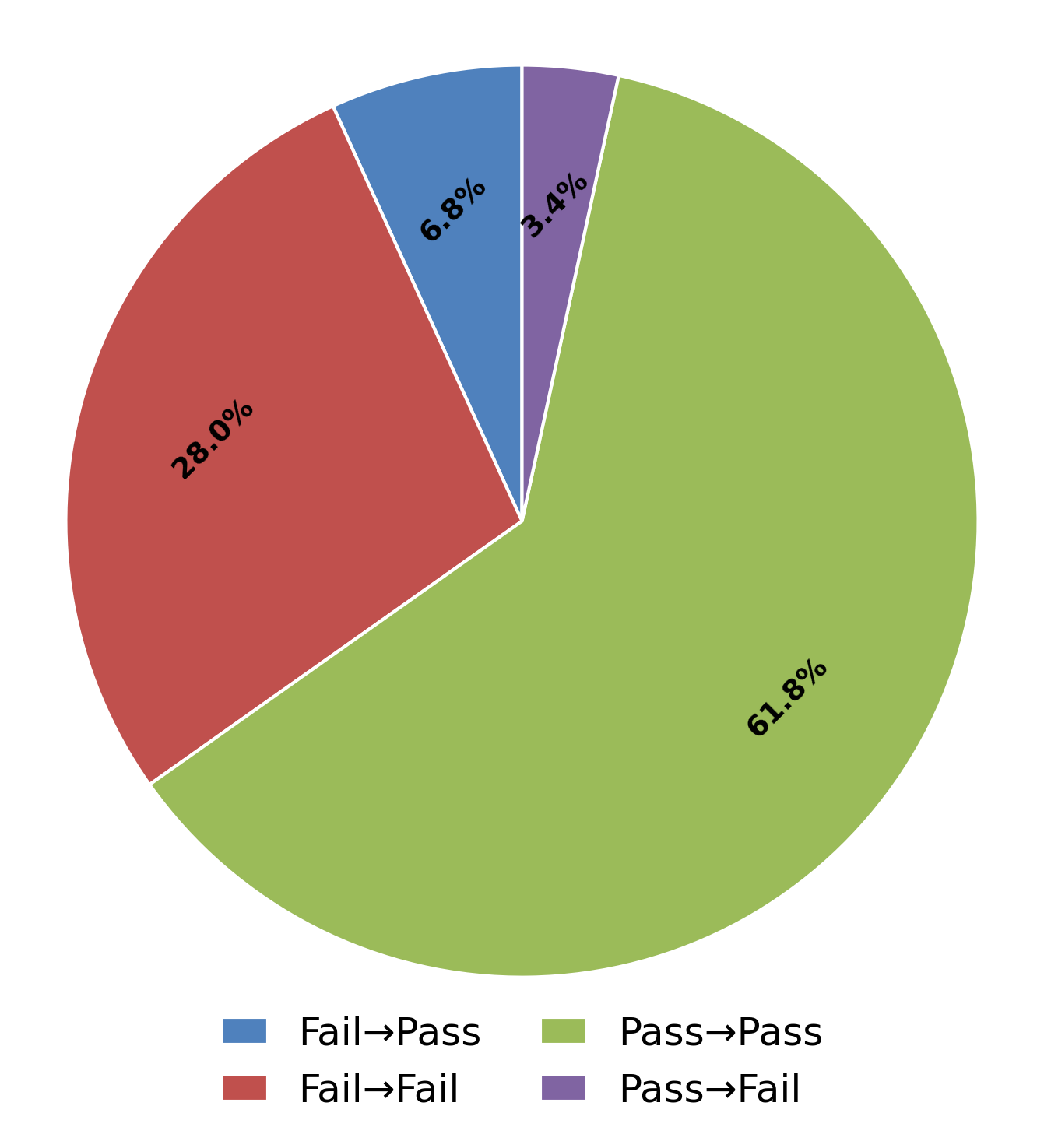}
\caption{Fine-granular outcome transitions from initial patch to SAGE. Majority of the outcomes stay consistent, but it flips from failure to success twice as often than pass-to-fail.}
\label{fig:outcome_transition}
\end{subfigure}
\caption{Distribution of how often SAGE-induced plans surface itself in the code changes between initial and SAGE-generated patches, indicating the correlation between the induced plan and the outcome changes. A SAGE-generated patch is considered to have a non-empty plan when at least one line in the code change can be attributed to at least one element in the plan.}
\end{figure}

Building on the qualitative observations from the case study, we now examine how the three reflective pillars—Analysis, Feedback, and Induced Plan—quantitatively surface within the final SAGE patches.
This section aims to characterize both the distribution of these pillars across code edits and the behavioral patterns that emerge when comparing original and SAGE-generated patches.

\noindent\textbf{Q1: How is plan attribution associated with outcome flips between the initial and SAGE patches?} 

We begin by examining how the induced plan surfaces across different outcome transitions and how its prominence relates to behavioral improvement or regression.
As shown in Figure~\ref{fig:plan_attribution_rate}, plan attributions are markedly stronger in SAGE-generated patches that \textit{change} the outcome, particularly those converting failures into passes as shown in Figure~\ref{fig:outcome_transition}.
This pattern suggests that successful recoveries are not random but anchored in explicit reflective grounding, whereas outcome-stable or regressive cases exhibit weaker or less coherent plan integration.
Although a small portion of cases exhibit \textit{pass→fail} transitions as expected, the beneficial flips (\textit{fail→pass}) occur more than twice as often (34 vs.\ 17 cases out of 500 on SWE-bench Verified), reinforcing that introspective planning predominantly facilitates corrective, rather than destabilizing, revisions.
Overall, these findings indicate that the strength of plan attribution serves as a reliable signal of effective self-guided reasoning—linking the agent’s reflective structure to tangible repair success.

\begin{figure}[tb]
\centering
\includegraphics[width=\linewidth]{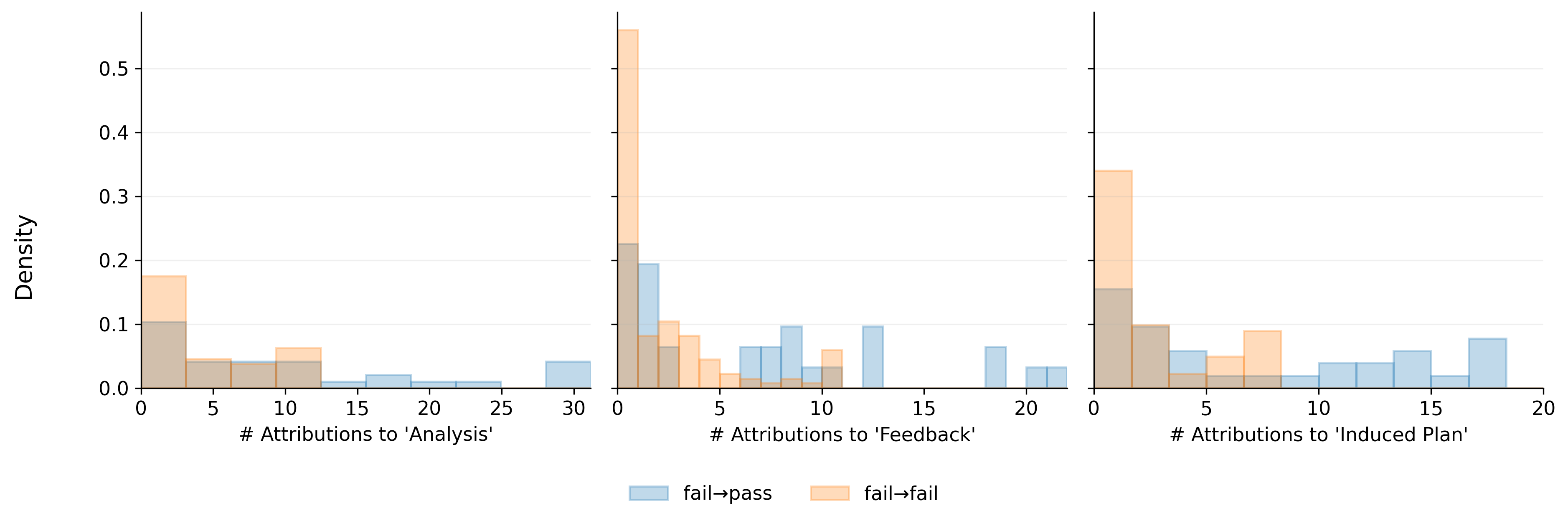}
\caption{Density of reflections of each pillar on the SAGE-generated final patches. It is obtained by the number of atomic rules in the SAGE plan grounded on the code change lines in the final SAGE patch. In summary, it shows how often the elements of the three main pillars of the overall plan surface in the actual code lines of the final patch.}
\label{fig:plan_pillar_density_fail2pass_fail2fail}
\end{figure}

\noindent\textbf{Q2: How are the finer-grained reflective pillars (Analysis, Feedback, Induced Plan) associated with changes in outcome?} 

Figure \ref{fig:plan_pillar_density_fail2pass_fail2fail} contrasts the density distributions of pillar attributions between \textit{fail→pass} and \textit{fail→fail} cases across the three reflective dimensions.
First, we observe that all three pillars surface more prominently in the \textit{fail→pass} patches, indicating that successful SAGE repairs tend to incorporate a richer trace of reflective reasoning, whereas failed ones remain sparsely attributed.
Second, the \textbf{Induced Plan} pillar exhibits the clearest separation in density, with its attribution counts concentrated toward higher values for \textit{fail→pass} cases, suggesting that the explicit plan abstraction provides a stronger grounding for corrective edits.
Third, the joint elevation across all pillars implies that SAGE’s introspective planning mechanism does not merely bias one type of reasoning, but collectively strengthens the agent’s capacity to link analysis, feedback, and planning signals to actionable code transformations.
Together, these patterns support the hypothesis that the presence and magnitude of plan-surfaced signals are predictive of whether self-guided reflection successfully converts failed trajectories into correct repairs.

\subsection{Impact of Bug Localization}

Bug localization is a crucial step for bug fixing tasks. Prior work~\cite{yang2024swe,xia2024agentless,liu2025empirical,chen-etal-2025-locagent} has shown that it contributes to a significant portion of failed repair cases. In this section, we investigate the effectiveness of providing additional information on bug localization for the agent. In particular, we evaluate GPT-5-mini (medium reasoning effort) on the full SWE-bench Verified dataset using the \texttt{mini-swe-agent} framework with the following setup: 1) Oracle -- where only the ground-truth localization is provided to the agent; 2) Mix -- ground-truth localization is provided but mixed with other top identified candidates from a variant of Swerank~\cite{reddy2025swerank}.  The mix setting is provided to mimic a more realistic setting as one would expect in practice.

As can be observed from Table \ref{tab:gpt5mini_localization}, the Oracle setting indeed greatly improves the performance of the agent, indicating the importance of bug localization. However, there is minimal improvement in the more realistic mixed setting. This suggests that noisy or uncertain localization cues do not substantially help the agent in deciding the location of the bug. These results indicate that high-precision localization signals can meaningfully enhance bug fixing, whereas ambiguous or partially correct ones offer limited benefit. 

\begin{table}[h] %
\centering

\begin{tabular}{lc}
\toprule
Setting & Resolved [\%] \\
\midrule
Baseline & 58.6 \\
Mix & 59.0 \\
Oracle & 64.4 \\
\bottomrule
\end{tabular}
\caption{Resolution rates of GPT-5-mini under different localization settings.}
\label{tab:gpt5mini_localization}
\end{table}

\subsection{Analysis of Git Usage}

\begin{table}[ht]
\centering

\resizebox{\columnwidth}{!}{%
    \begin{tabular}{l c rrr} 
    \toprule
     & {\textbf{Potentially Unsafe}} & \multicolumn{3}{c}{\textbf{Relative Frequency [\%]}} \\
    \cmidrule(lr){3-5}
    \textbf{Agent} & {\textbf{Actions [\%] }} & \textbf{\texttt{git log}} & \textbf{\texttt{git show}} & \textbf{\texttt{git diff}} \\
    \midrule
    GPT-5 & 0.000 & 0.0 & 0.0 & 0.0 \\
    GPT-5 (high) & 0.016 & 0.0 & 0.0 & 100.0 \\
    GPT-5 (high) + Gemini 2.5 Plan & 0.004 & 0.0 & 0.0 & 100.0 \\
    Claude Sonnet 4 & 0.004 & 50.0 & 50.0 & 0.0 \\
    Claude Sonnet 4.5 & 0.389 & 54.0 & 46.0 & 0.0 \\
    \bottomrule
    \end{tabular}
}
\caption{Frequency of potentially unsafe \texttt{git} commands, which could leak future git history in unsafeguarded environments. We report the total percentage of such actions and the relative frequency of specific commands. \textbf{Our experiments are run using Docker images that prevent such leakage.}}
\label{tab:git-analysis}
\end{table}

Recent findings have exposed models using git operations to view future states of the repository during rollouts, thereby leaking the gold solution\footnote{\url{https://github.com/SWE-bench/SWE-bench/issues/465/}}. To ensure the integrity of our results, we use the latest SWE-bench Verified Docker images. These images have removed the future git history, thus preventing the model from accessing information about the merged PR for the given issue. While this safeguards the model from accessing information about the solution, we observed that agents still execute git commands that could leak data in unsafeguarded containers (\textit{e.g.}, \inlinebox{\texttt{git log -{}-all}}). To understand this behavior and inform future benchmark design, we analyzed the git actions made by agents that could reveal future information in environments without these safeguards (Table~\ref{tab:git-analysis}). GPT-5 and Claude Sonnet 4 rarely made such git calls (< $0.02\%$ of total actions), with GPT-5 favoring \inlinebox{\texttt{git diff}} and Claude Sonnet 4 favoring \inlinebox{\texttt{git log}} and \inlinebox{\texttt{git show}} commands. Notably, Claude Sonnet 4.5, the best performing model in our experiments, is also the model that made the most git calls that could potentially leak future information ($0.39\%$ of total actions) in environments without such safeguards. Manual inspection of these trajectories revealed two key findings. First, most operations were targeted, not exploratory; they involved specific commit lookups (\textit{e.g.}, \inlinebox{\texttt{git show <hash>}}) or included a \inlinebox{\texttt{grep}} operation that limited information retrieval. Second, these commands are typically grounded in the problem context: the commit hash or \inlinebox{\texttt{grep}} pattern originated from the issue description in $79.4\%$ of cases. These grounded calls were most frequent in Django instances, whose issue statements often reference specific commits or ticket numbers. However, even these seemingly-benign, grounded operations pose a leakage risk in environments without relevant safeguards. For example, a \inlinebox{\texttt{grep}} pattern from the issue could unintentionally match keywords in the solution commit, or a command like \inlinebox{\texttt{git log -{}-since="<date>"}} could leak repository information that extends beyond the solution's merge date. We leave the design of more robust safeguards for LLM-based software engineering evaluations as future work.

\subsection{Connection to Other Reinforcement Learning Formulations}
\subsubsection{Semi-Markov Decision Process}
SAGE can also be viewed from an \textit{option} \cite{sutton1999between} perspective. Recall, an option $o \in \mathcal{O}$ is a tuple
\[
o \;=\; \big(\mathcal{I}_o,\; \mu_o,\; \beta_o\big),
\]
where $\mathcal{I}_o \subseteq \mathcal{S}$ is the initial set such that option $o$ may start only in $s\in\mathcal{I}_o$. $\mu_o$ is the intra-option policy, used when option $o$ is selected. $\beta_o$ is the termination condition that indicates the probability $o$ terminates upon arrival in state $s$.
An option is a temporal abstraction over actions which represents a high-level ``summary'' of the corresponding action sequence.

In this framework, the plan abstraction and augmented execution phase can be viewed as an option, where the option initial set $\mathcal{I}_o$ is the  explore agent $\mathcal{A}_\theta$ terminal state. A plan $\psi$ is then induced following Equation \ref{eq:plan_induce}. The option policy $\mu_o$ is then augmented with the generated plan $\psi$, which is equivalent to $\pi_{\theta}^+$.

\subsubsection{Bayesian Reinforcement Learning}
We can also view SAGE through the lens of Bayesian Reinforcement Learning (RL) \cite{bellman2003adaptive, martin1967bayesian, ghavamzadeh2015bayesian}. Bayesian Reinforcement Learning views the underlying MDP $\mathcal{M}$ as a random variable and seeks to maximize the expected reward under the policy and the posterior distribution of $\mathcal{M}$. Thus, Bayesian RL methods maintain a posterior $p(\mathcal{M} \mid \mathcal{H}_t)$, given the history $\mathcal{H}_t$, and learn a policy that maximizes expected reward under this posterior. Concretely, SAGE can be seen as a special case of posterior sampling for reinforcement learning (PSRL) \cite{strens2000bayesian} limited to 2 episodes. For each episode, PSRL has three steps: (1) sample an MDP from the posterior, $\mathcal{M} \sim p(\mathcal{M} \mid \mathcal{H})$, (2) compute and execute the optimal policy $\pi^*$ for the given MDP $\mathcal{M}$, (3) update the posterior given the latest episode. In SAGE, the initial rollout constitutes the first episode. Here, the initial posterior over $\mathcal{M}$ is implicitly defined by the LLM's prior, and the optimal policy is approximated by the actor $\pi_{\theta}$. Following this episode, the posterior is updated in-context via the plan abstraction step, incorporating information from the new trajectory. The final rollout in SAGE can be viewed as the second episode, where SAGE samples a plan from the LLM (analogous to step (1): sampling an MDP $\mathcal{M}$) and then executes the policy $\pi_{\theta}$ conditioned on this plan (analogous to step (2)). However, SAGE deviates from PSRL in step (3) as SAGE performs only one posterior update (after episode 1) and does not update the posterior after the second episode. We leave multi-episode extensions of SAGE as future work.

\section{Related Work}
As large language models (LLMs) demonstrate strong capabilities in general coding~\cite{Chen2021Evaluating,Nijkamp2022CodeGen,Li2023StarCoder,anthropic2025sonnet4.5,openai2025gpt5} and software engineering tasks~\cite{xia2024agentless,moatless-tools,zhang2024autocoderover,zhang2023repocoder,kovacic2025refactoring}, attention has increasingly shifted to evaluating and improving them for complex, multi-step agentic tasks~\cite{yao2022webshop,osworld_verified,zhou2024webarena,koh2024visualwebarena,mundler2024swt}.
For SWE-Bench~\cite{jimenez2024swebench}, steady progress has been made from various angles: (1) stronger base models, (2) incorporating effective tools~\cite{wang2025openhands,moatless-tools,refact2025sota}, and (3) intelligently scaling test-time compute~\cite{traeresearchteam2025traeagent}, among others.
In this work, we introduce SAGE, a novel method that operates orthogonally to these research directions. We show that SAGE exhibits consistent improvements across models and agent frameworks, demonstrating its broad compatibility and additive value.

Another critical line of work focuses on agent self-improvement and iterative refinement. Instead of a single-shot generation, agents are designed to reflect on their own outputs, learn from failures, and progressively correct their solutions. For instance, agents may employ an iterative "critique-and-refine" loop, using a sub-model to analyze their own patch for flaws before generating an improved version~\cite{refact2025sota,shinn2023reflexion,le2024indict}. Others use tree search techniques like MCTS to explore and evaluate different patch candidates, refining the solution incrementally~\cite{hu2025aprmcts,antoniades2025swesearch,li2024codetree,ma2025alibaba}. A different line of work focuses on learning from experience, distilling knowledge from annotated trajectories into an ``experience bank'' to guide future attempts~\cite{chen2025swe}. More advanced methods even allow the agent to optimize its own internal code or prompts, growing a tree of self-modifications to discover a more effective agent design~\cite{zhang2025darwin,novikov2025alphaevolve}.
SAGE shares similarities with previous methods, but demonstrates that a simple trajectory-level experience reflection as a plan is effective.
 
Various works rely on some forms of planning, or ``scaffold,'' by informing the agent to follow its own multi-step plan~\cite{yang2024swe,wang2025openhands,huang2022inner,hao2023reasoning}. Agents following scaffolds retain the flexibility to adapt their actions.
For SWE-Bench, a typical trajectory involves investigating the repository, writing a script to reproduce the bug, and then iteratively proposing, testing, and debugging a fix.
More advanced agents can also perform explicit plan revision, using specialized tools or reflection steps to analyze failures and generate a new plan of action mid-task~\cite{refact2025sota}.
SAGE starts with a scaffold but augments the prompt with feedback and a revision plan in the final execution.

\section{Conclusion}
We propose Self-Abstraction from Grounded Experience (SAGE) -- a test-time adaptation framework that allows agent to learn and improve based on their own experience. SAGE implements a structured cycle of exploration, plan abstract, and re-execution to learn from complete rollouts. This process uncovers grounded and informative guidance about the environment and task, which is then used to guide the subsequent execution. As a drop-in method, SAGE is easily applied to any agent framework, and we demonstrate its effectiveness across different LLM backbones and agent architectures. Furthermore, ablation and analysis confirm that the induced grounded plan is crucial for realizing subsequent performance improvements.

\newpage
\bibliographystyle{plain}
\bibliography{reference}

\begin{thebibliography}{10}

\bibitem{anthropic2025sonnet4}
Anthropic.
\newblock Introducing claude sonnet 4.
\newblock Anthropic News, May 2025.

\bibitem{anthropic2025sonnet4.5}
Anthropic.
\newblock Introducing claude sonnet 4.5.
\newblock Anthropic News, Sep 2025.

\bibitem{antoniades2025swesearch}
Antonis Antoniades, Albert {\"O}rwall, Kexun Zhang, Yuxi Xie, Anirudh Goyal, and William~Yang Wang.
\newblock {SWE}-search: Enhancing software agents with monte carlo tree search and iterative refinement.
\newblock In {\em The Thirteenth International Conference on Learning Representations}, 2025.

\bibitem{bellman2003adaptive}
Richard Bellman and Robert Kalaba.
\newblock On adaptive control processes.
\newblock {\em IRE Transactions on Automatic Control}, 4(2):1--9, 2003.

\bibitem{Chen2021Evaluating}
Mark Chen, Jerry Tworek, Heewoo Jun, Qiming Yuan, Henrique Pond{\'e}, Jared Kaplan, Harrison Edwards, Yura Burda, Nicholas Joseph, Greg Brockman, Alex Ray, Raul Puri, Gretchen Krueger, Michael Petrov, Heidy Khlaaf, Girish Sastry, Pamela Mishkin, Brooke Chan, Scott Gray, Nick Ryder, Mikhail Pavlov, Alethea Power, Lukasz Kaiser, Mo~Bavarian, Clemens Winter, Phil Tillet, Felipe~Petroski Such, David~W. Cummings, Matthias Plappert, Fotios Chantzis, Elizabeth Barnes, Ariel Herbert-Voss, William~H. Guss, Alex Nichol, Igor Babuschkin, Suchir Balaji, Shantanu Jain, Andrew Carr, Jan Leike, Josh Achiam, Vedant Misra, Evan Morikawa, Alec Radford, Matthew~M. Knight, Miles Brundage, Mira Murati, Katie Mayer, Peter Welinder, Bob McGrew, Dario Amodei, Sam McCandlish, Ilya Sutskever, and Wojciech Zaremba.
\newblock Evaluating large language models trained on code.
\newblock {\em ArXiv}, abs/2107.03374, 2021.

\bibitem{chen2025swe}
Silin Chen, Shaoxin Lin, Xiaodong Gu, Yuling Shi, Heng Lian, Longfei Yun, Dong Chen, Weiguo Sun, Lin Cao, and Qianxiang Wang.
\newblock Swe-exp: Experience-driven software issue resolution.
\newblock {\em arXiv preprint arXiv:2507.23361}, 2025.

\bibitem{chen-etal-2025-locagent}
Zhaoling Chen, Robert Tang, Gangda Deng, Fang Wu, Jialong Wu, Zhiwei Jiang, Viktor Prasanna, Arman Cohan, and Xingyao Wang.
\newblock {L}oc{A}gent: Graph-guided {LLM} agents for code localization.
\newblock In Wanxiang Che, Joyce Nabende, Ekaterina Shutova, and Mohammad~Taher Pilehvar, editors, {\em Proceedings of the 63rd Annual Meeting of the Association for Computational Linguistics (Volume 1: Long Papers)}, pages 8697--8727, Vienna, Austria, July 2025. Association for Computational Linguistics.

\bibitem{chowdhury2024swebenchverified}
Neil Chowdhury, James Aung, Chan~Jun Shern, Oliver Jaffe, Dane Sherburn, Giulio Starace, Evan Mays, Rachel Dias, Marwan Aljubeh, Mia Glaese, Carlos~E. Jimenez, John Yang, Leyton Ho, Tejal Patwardhan, Kevin Liu, and Aleksander Madry.
\newblock Introducing {SWE}-bench verified.
\newblock OpenAI Blog, 2024.

\bibitem{ghavamzadeh2015bayesian}
Mohammad Ghavamzadeh, Shie Mannor, Joelle Pineau, Aviv Tamar, et~al.
\newblock Bayesian reinforcement learning: A survey.
\newblock {\em Foundations and Trends{\textregistered} in Machine Learning}, 8(5-6):359--483, 2015.

\bibitem{hao2023reasoning}
Shibo Hao, Yi~Gu, Haodi Ma, Joshua~Jiahua Hong, Zhen Wang, Daisy~Zhe Wang, and Zhiting Hu.
\newblock Reasoning with language model is planning with world model.
\newblock {\em arXiv preprint arXiv:2305.14992}, 2023.

\bibitem{hu2025aprmcts}
Haichuan Hu, Congqing He, Hao Zhang, Xiaochen Xie, and Quanjun Zhang.
\newblock Aprmcts: Improving llm-based automated program repair with iterative tree search.
\newblock {\em arXiv preprint arXiv:2507.01827}, 2025.

\bibitem{huang2022inner}
Wenlong Huang, Fei Xia, Ted Xiao, Harris Chan, Jacky Liang, Pete Florence, Andy Zeng, Jonathan Tompson, Igor Mordatch, Yevgen Chebotar, et~al.
\newblock Inner monologue: Embodied reasoning through planning with language models.
\newblock {\em arXiv preprint arXiv:2207.05608}, 2022.

\bibitem{jimenez2024swebench}
Carlos~E Jimenez, John Yang, Alexander Wettig, Shunyu Yao, Kexin Pei, Ofir Press, and Karthik~R Narasimhan.
\newblock {SWE}-bench: Can language models resolve real-world github issues?
\newblock In {\em The Twelfth International Conference on Learning Representations}, 2024.

\bibitem{koh2024visualwebarena}
Jing~Yu Koh, Robert Lo, Lawrence Jang, Vikram Duvvur, Ming Lim, Po-Yu Huang, Graham Neubig, Shuyan Zhou, Russ Salakhutdinov, and Daniel Fried.
\newblock Visualwebarena: Evaluating multimodal agents on realistic visual web tasks.
\newblock In {\em Proceedings of the 62nd Annual Meeting of the Association for Computational Linguistics (Volume 1: Long Papers)}, pages 881--905, 2024.

\bibitem{kovacic2025refactoring}
Ziga Kovacic, Justin~T Chiu, Celine Lee, Wenting Zhao, and Kevin Ellis.
\newblock Refactoring codebases through library design.
\newblock {\em arXiv preprint arXiv:2506.11058}, 2025.

\bibitem{le2024indict}
Hung Le, Doyen Sahoo, Yingbo Zhou, Caiming Xiong, and Silvio Savarese.
\newblock Indict: Code generation with internal dialogues of critiques for both security and helpfulness.
\newblock {\em Advances in Neural Information Processing Systems}, 37:85546--85582, 2024.

\bibitem{li2024evocodebench}
Jia Li, Ge~Li, Xuanming Zhang, Yunfei Zhao, Yihong Dong, Zhi Jin, Binhua Li, Fei Huang, and Yongbin Li.
\newblock Evocodebench: An evolving code generation benchmark with domain-specific evaluations.
\newblock {\em Advances in Neural Information Processing Systems}, 37:57619--57641, 2024.

\bibitem{li2024codetree}
Jierui Li, Hung Le, Yingbo Zhou, Caiming Xiong, Silvio Savarese, and Doyen Sahoo.
\newblock Codetree: Agent-guided tree search for code generation with large language models.
\newblock {\em arXiv preprint arXiv:2411.04329}, 2024.

\bibitem{Li2023StarCoder}
Raymond Li, Loubna~Ben allal, Yangtian Zi, Niklas Muennighoff, Denis Kocetkov, Chenghao Mou, Marc Marone, Christopher Akiki, Jia LI, Jenny Chim, Qian Liu, Evgenii Zheltonozhskii, Terry~Yue Zhuo, Thomas Wang, Olivier Dehaene, Joel Lamy-Poirier, Joao Monteiro, Nicolas Gontier, Ming-Ho Yee, Logesh~Kumar Umapathi, Jian Zhu, Ben Lipkin, Muhtasham Oblokulov, Zhiruo Wang, Rudra Murthy, Jason~T Stillerman, Siva~Sankalp Patel, Dmitry Abulkhanov, Marco Zocca, Manan Dey, Zhihan Zhang, Urvashi Bhattacharyya, Wenhao Yu, Sasha Luccioni, Paulo Villegas, Fedor Zhdanov, Tony Lee, Nadav Timor, Jennifer Ding, Claire~S Schlesinger, Hailey Schoelkopf, Jan Ebert, Tri Dao, Mayank Mishra, Alex Gu, Carolyn~Jane Anderson, Brendan Dolan-Gavitt, Danish Contractor, Siva Reddy, Daniel Fried, Dzmitry Bahdanau, Yacine Jernite, Carlos~Mu{\~n}oz Ferrandis, Sean Hughes, Thomas Wolf, Arjun Guha, Leandro~Von Werra, and Harm de~Vries.
\newblock Starcoder: may the source be with you!
\newblock {\em Transactions on Machine Learning Research}, 2023.

\bibitem{liu-etal-2025-repodebug}
Jingjing Liu, Zeming Liu, Zihao Cheng, Mengliang He, Xiaoming Shi, Yuhang Guo, Xiangrong Zhu, Yuanfang Guo, Yunhong Wang, and Haifeng Wang.
\newblock {R}epo{D}ebug: Repository-level multi-task and multi-language debugging evaluation of large language models.
\newblock In Christos Christodoulopoulos, Tanmoy Chakraborty, Carolyn Rose, and Violet Peng, editors, {\em Findings of the Association for Computational Linguistics: EMNLP 2025}, pages 23784--23813, Suzhou, China, November 2025. Association for Computational Linguistics.

\bibitem{liu2025empirical}
Simiao Liu, Fang Liu, Liehao Li, Xin Tan, Yinghao Zhu, Xiaoli Lian, and Li~Zhang.
\newblock An empirical study on failures in automated issue solving.
\newblock {\em arXiv preprint arXiv:2509.13941}, 2025.

\bibitem{liu2023repobench}
Tianyang Liu, Canwen Xu, and Julian McAuley.
\newblock Repobench: Benchmarking repository-level code auto-completion systems.
\newblock In {\em The Twelfth International Conference on Learning Representations}, 2024.

\bibitem{ma2025alibaba}
Yingwei Ma, Qingping Yang, Rongyu Cao, Binhua Li, Fei Huang, and Yongbin Li.
\newblock Alibaba lingmaagent: Improving automated issue resolution via comprehensive repository exploration.
\newblock In {\em Proceedings of the 33rd ACM International Conference on the Foundations of Software Engineering}, pages 238--249, 2025.

\bibitem{madaan2023self}
Aman Madaan, Niket Tandon, Prakhar Gupta, Skyler Hallinan, Luyu Gao, Sarah Wiegreffe, Uri Alon, Nouha Dziri, Shrimai Prabhumoye, Yiming Yang, et~al.
\newblock Self-refine: Iterative refinement with self-feedback.
\newblock {\em Advances in Neural Information Processing Systems}, 36:46534--46594, 2023.

\bibitem{martin1967bayesian}
James~John Martin.
\newblock {\em Bayesian decision problems and Markov chains}.
\newblock John Wiley \& Sons, 1967.

\bibitem{mundler2024swt}
Niels M{\"u}ndler, Mark M{\"u}ller, Jingxuan He, and Martin Vechev.
\newblock Swt-bench: Testing and validating real-world bug-fixes with code agents.
\newblock {\em Advances in Neural Information Processing Systems}, 37:81857--81887, 2024.

\bibitem{Nijkamp2022CodeGen}
Erik Nijkamp, Bo~Pang, Hiroaki Hayashi, Lifu Tu, Haiquan Wang, Yingbo Zhou, Silvio Savarese, and Caiming Xiong.
\newblock Codegen: An open large language model for code with multi-turn program synthesis.
\newblock In {\em International Conference on Learning Representations}, 2022.

\bibitem{novikov2025alphaevolve}
Alexander Novikov, Ng{\^a}n V{\~u}, Marvin Eisenberger, Emilien Dupont, Po-Sen Huang, Adam~Zsolt Wagner, Sergey Shirobokov, Borislav Kozlovskii, Francisco~JR Ruiz, Abbas Mehrabian, et~al.
\newblock Alphaevolve: A coding agent for scientific and algorithmic discovery.
\newblock {\em arXiv preprint arXiv:2506.13131}, 2025.

\bibitem{openai2025gpt5}
OpenAI.
\newblock Introducing gpt-5.
\newblock OpenAI Blog, Aug 2025.

\bibitem{moatless-tools}
Albert {\"O}rwall.
\newblock moatless-tools: A framework for experimenting with llms for code editing.
\newblock \url{https://github.com/aorwall/moatless-tools}, 2024.

\bibitem{panickssery2024llm}
Arjun Panickssery, Samuel Bowman, and Shi Feng.
\newblock Llm evaluators recognize and favor their own generations.
\newblock {\em Advances in Neural Information Processing Systems}, 37:68772--68802, 2024.

\bibitem{reddy2025swerank}
Revanth~Gangi Reddy, Tarun Suresh, JaeHyeok Doo, Ye~Liu, Xuan~Phi Nguyen, Yingbo Zhou, Semih Yavuz, Caiming Xiong, Heng Ji, and Shafiq Joty.
\newblock Swerank: Software issue localization with code ranking.
\newblock {\em arXiv preprint arXiv:2505.07849}, 2025.

\bibitem{shinn2023reflexion}
Noah Shinn, Federico Cassano, Ashwin Gopinath, Karthik Narasimhan, and Shunyu Yao.
\newblock Reflexion: Language agents with verbal reinforcement learning.
\newblock {\em Advances in Neural Information Processing Systems}, 36:8634--8652, 2023.

\bibitem{strens2000bayesian}
Malcolm Strens.
\newblock A bayesian framework for reinforcement learning.
\newblock In {\em ICML}, volume 2000, pages 943--950, 2000.

\bibitem{sutton1999between}
Richard~S Sutton, Doina Precup, and Satinder Singh.
\newblock Between mdps and semi-mdps: A framework for temporal abstraction in reinforcement learning.
\newblock {\em Artificial intelligence}, 112(1-2):181--211, 1999.

\bibitem{Gemini2.5Pro2025}
Gemini Team and Google.
\newblock Gemini 2.5: Pushing the frontier with advanced reasoning, multimodality, long context, and next generation agentic capabilities.
\newblock {\em arXiv preprint arXiv:2507.06261}, 2025.

\bibitem{traeresearchteam2025traeagent}
Trae~Research Team, Pengfei Gao, Zhao Tian, Xiangxin Meng, Xinchen Wang, Ruida Hu, Yuanan Xiao, Yizhou Liu, Zhao Zhang, Junjie Chen, Cuiyun Gao, Yun Lin, Yingfei Xiong, Chao Peng, and Xia Liu.
\newblock Trae agent: An llm-based agent for software engineering with test-time scaling.
\newblock 2025.

\bibitem{refact2025sota}
Sergey Vakhreev and {Refact AI Team}.
\newblock Open-source sota on swe-bench verified: Refact ai.
\newblock Refact AI Blog, 2025.

\bibitem{wang2025openhands}
Xingyao Wang, Boxuan Li, Yufan Song, Frank~F. Xu, Xiangru Tang, Mingchen Zhuge, Jiayi Pan, Yueqi Song, Bowen Li, Jaskirat Singh, Hoang~H. Tran, Fuqiang Li, Ren Ma, Mingzhang Zheng, Bill Qian, Yanjun Shao, Niklas Muennighoff, Yizhe Zhang, Binyuan Hui, Junyang Lin, Robert Brennan, Hao Peng, Heng Ji, and Graham Neubig.
\newblock Openhands: An open platform for {AI} software developers as generalist agents.
\newblock In {\em The Thirteenth International Conference on Learning Representations}, 2025.

\bibitem{xia2024agentless}
Chunqiu~Steven Xia, Yinlin Deng, Soren Dunn, and Lingming Zhang.
\newblock Agentless: Demystifying llm-based software engineering agents.
\newblock {\em arXiv preprint arXiv:2407.01489}, 2024.

\bibitem{osworld_verified}
Tianbao Xie, Mengqi Yuan, Danyang Zhang, Xinzhuang Xiong, Zhennan Shen, Zilong Zhou, Xinyuan Wang, Yanxu Chen, Jiaqi Deng, Junda Chen, Bowen Wang, Haoyuan Wu, Jixuan Chen, Junli Wang, Dunjie Lu, Hao Hu, and Tao Yu.
\newblock Introducing osworld-verified.
\newblock {\em xlang.ai}, July 2025.

\bibitem{yang2024swe}
John Yang, Carlos~E Jimenez, Alexander Wettig, Kilian Lieret, Shunyu Yao, Karthik Narasimhan, and Ofir Press.
\newblock Swe-agent: Agent-computer interfaces enable automated software engineering.
\newblock {\em Advances in Neural Information Processing Systems}, 37:50528--50652, 2024.

\bibitem{yao2022webshop}
Shunyu Yao, Howard Chen, John Yang, and Karthik Narasimhan.
\newblock Webshop: Towards scalable real-world web interaction with grounded language agents.
\newblock In S.~Koyejo, S.~Mohamed, A.~Agarwal, D.~Belgrave, K.~Cho, and A.~Oh, editors, {\em Advances in Neural Information Processing Systems}, volume~35, pages 20744--20757. Curran Associates, Inc., 2022.

\bibitem{zhang2023repocoder}
Fengji Zhang, Bei Chen, Yue Zhang, Jacky Keung, Jin Liu, Daoguang Zan, Yi~Mao, Jian-Guang Lou, and Weizhu Chen.
\newblock Repocoder: Repository-level code completion through iterative retrieval and generation.
\newblock {\em arXiv preprint arXiv:2303.12570}, 2023.

\bibitem{zhang2025darwin}
Jenny Zhang, Shengran Hu, Cong Lu, Robert Lange, and Jeff Clune.
\newblock Darwin godel machine: Open-ended evolution of self-improving agents.
\newblock {\em arXiv preprint arXiv:2505.22954}, 2025.

\bibitem{zhang2024diversity}
Kexun Zhang, Weiran Yao, Zuxin Liu, Yihao Feng, Zhiwei Liu, Rithesh Murthy, Tian Lan, Lei Li, Renze Lou, Jiacheng Xu, et~al.
\newblock Diversity empowers intelligence: Integrating expertise of software engineering agents.
\newblock In {\em The Thirteenth International Conference on Learning Representations}, 2025.

\bibitem{zhang2024autocoderover}
Yuntong Zhang, Haifeng Ruan, Zhiyu Fan, and Abhik Roychoudhury.
\newblock Autocoderover: Autonomous program improvement.
\newblock In {\em Proceedings of the 33rd ACM SIGSOFT International Symposium on Software Testing and Analysis}, pages 1592--1604, 2024.

\bibitem{zheng2023judging}
Lianmin Zheng, Wei-Lin Chiang, Ying Sheng, Siyuan Zhuang, Zhanghao Wu, Yonghao Zhuang, Zi~Lin, Zhuohan Li, Dacheng Li, Eric Xing, et~al.
\newblock Judging llm-as-a-judge with mt-bench and chatbot arena.
\newblock {\em Advances in neural information processing systems}, 36:46595--46623, 2023.

\bibitem{zhou2023language}
Andy Zhou, Kai Yan, Michal Shlapentokh-Rothman, Haohan Wang, and Yu-Xiong Wang.
\newblock Language agent tree search unifies reasoning acting and planning in language models.
\newblock {\em arXiv preprint arXiv:2310.04406}, 2023.

\bibitem{zhou2024webarena}
Shuyan Zhou, Frank~F. Xu, Hao Zhu, Xuhui Zhou, Robert Lo, Abishek Sridhar, Xianyi Cheng, Tianyue Ou, Yonatan Bisk, Daniel Fried, Uri Alon, and Graham Neubig.
\newblock Webarena: A realistic web environment for building autonomous agents.
\newblock In {\em The Twelfth International Conference on Learning Representations}, 2024.

\end{thebibliography}

\newpage
\appendix

\section{Prompts}
\label{app:prompts}
We list the prompt templates used for plan and plan-augmented execution agents below.
Common formatting instruction sections are noted and omitted as they are directly imported from the original mini-swe-agent.

\begin{tcolorbox}[title=Plan Abstraction (system),float, floatplacement=ht]
{\tiny
\begin{Verbatim}[breaklines,breaksymbol=]
You are an expert Principal Software Engineer acting as an automated reviewer. 
Your primary goal is to analyze the work of an autonomous AI software agent that attempted to fix a bug. 
You must assess its solution and then provide a customized high-level plan based on the analysis.
You can interact multiple times with a computer shell to solve programming tasks.
Your response must contain exactly ONE bash code block with ONE command (or commands connected with && or ||).

You will be provided with two pieces of information:
The original bug report in <issue_description> section, which describes the issue.
The complete hisotry is presented in the <resolution_attempt> section. 
The <resolution_attempt> section contains the complete trajectory that includes the thought, command, and observation during its attempt to fix the bug from another agent.
Your review must be based solely on the information provided in the trajectory. Do not make assumptions about code or context not present in the logs.

# Core Task: Review and Revise
Your review process must follow these three steps:
Reconstruct the Solver's Strategy: 
1. Carefully analyze the entire action trajectory from start to finish. From the sequence of actions, infer the solver agent's implicit high-level plan. What was its hypothesis about the bug's root cause? What was its intended solution path?
2. Evaluate the Plan and Execution: Critically evaluate the solver's strategy and its implementation.
3. Plan Evaluation: Was the high-level plan logical and sound? Was it an efficient way to approach this specific bug, or was there a more direct method?
4. Execution Evaluation: Did the agent's actions effectively implement its plan? Did it correctly diagnose the issue? Is the final code change correct and does it fully resolve the bug described in the report?

Deliver Your Complete Customized Plan: Based on your analysis and repository information collected from the preivous attempt, provide a complete and customized high level plan for the given problem in the context of the given git repository. 
IMPORTANT: the plan should be complete, keep the plan at high level, relevent to the problem presented, and do not enumerate all of the details

<instructions>

[FORMATTING INSTRUCTIONS FROM MINI-SWE-AGENT ABBREVIATED]

## Submission
When you've completed your work (reading, editing, testing), and cannot make further progress add the following in your THOUGHT section:
<analysis>
### 1. Inferred High-Level Plan
[Based on the trajectory, describe the high-level plan you believe the solver agent was following. Use a bulleted or numbered list.]

### 2. Evaluation of the Plan's Logic
[Critique the plan itself, *before* considering its implementation. Was this a sensible strategy for the given bug? Was its initial hypothesis correct? Mention any conceptual flaws.]

### 3. Review of Implementation and Final Code
[Analyze how the agent executed its plan. Point out key steps, both successful and unsuccessful. State clearly whether the final code modification is correct and if it solves the original problem.]
</analysis>

<feedback>
## Corrective Feedback
</feedback>

<new_plan>
### Customized High-Level Plan
[Provide a customized complete high-level plan as a clear, actionable list.]
</new_plan>

### Explanation for Revision
[Explain why your revised plan is superior. Directly address the logical errors or flawed assumptions made by the solver agent and how your plan avoids them.]

Then, issue exactly the following command:

```bash
echo COMPLETE_TASK_AND_SUBMIT_FINAL_OUTPUT && git add -A && git diff --cached
```

This command will submit your work.
You cannot continue working (reading, editing, testing) in any way on this task after submitting.
</instructions>
\end{Verbatim}
}
\end{tcolorbox}

\begin{tcolorbox}[title=Plan Abstraction (user),float, floatplacement=ht]
{\tiny
\begin{Verbatim}[breaklines,breaksymbol=]
Your primary goal is to analyze the work of an autonomous AI software agent that attempted to fix a bug. 
You must assess the correctness of its solution and provide constructive, high-level feedback if it failed.

IMPORTANT: This is an interactive process where you will think and issue ONE command, see its result, then think and issue your next command.
<issue_description>
{{task}}
</issue_description>

<resolution_attempt>
{%
{%
Observation: {{item.content}}
{%
Action: {{item.content}}
{%
{%
</resolution_attempt>
\end{Verbatim}
}
\end{tcolorbox}

\newpage
\begin{tcolorbox}[title=Plan-augmented Execution (user),float, floatplacement=ht]
{\tiny
\begin{Verbatim}[breaklines,breaksymbol=]
<pr_description>
Consider the following PR description:
{{task}}
</pr_description>
{%

<previous_attempt>
##Analysis:
{{plan.analysis}}
##Feedback:
{{plan.feedback}}
##New Plan:
{{plan.new_plan}}
</previous_attempt> 

{%
<instructions>
# Task Instructions

## Overview
You're a Principle software engineer interacting continuously with a computer by submitting commands.
You'll be helping implement necessary changes to meet requirements in the PR description.
Your task is specifically to make changes to non-test files in the current directory in order to fix the issue described in the PR description in a way that is general and consistent with the codebase.
{%
You are also provided with an analysis of previous attempt by another engineer to solve this PR in the <previous_attempt> section.
In the <previous_attempt> section, the previous plan along with its analysis, feedback and a new plan are provided.
Make sure to consider <new_plan> along with the rational and feedabck in resolving the issue described in the <pr_description> section
{%
IMPORTANT: This is an interactive process where you will think and issue ONE command, see its result, then think and issue your next command.

[FORMATTING INSTRUCTIONS FROM MINI-SWE-AGENT ABBREVIATED]

\end{Verbatim}
}
\end{tcolorbox}

\end{document}